\pdfoutput=1

\documentclass[11pt]{article}

\usepackage{acl}

\usepackage{times}
\usepackage{latexsym}
\usepackage{graphicx}
\usepackage{bm}
\usepackage{verbatimbox}
\usepackage{tabularx}
\usepackage{booktabs}
\usepackage{amsmath}
\usepackage{multirow}
\usepackage{paralist}
\usepackage{enumitem}
\usepackage{xcolor}
\usepackage{epigraph}
\usepackage{amsfonts}

\usepackage[T1]{fontenc}

\usepackage[utf8]{inputenc}

\usepackage{microtype}

\title{Ask to Understand: Question Generation for Multi-hop Question Answering}

\author{Jiawei Li, Mucheng Ren, Yang Gao\thanks{~~Corresponding author.}, Yizhe Yang \\
        School of Computer Science and Technology,\\
        Beijing Institute of Technology, Beijing, China \\
        Beijing Engineering Research Center of High Volume Language Information \\
        Processing and Cloud Computing Applications, Beijing, China\\
        \texttt{\{jwli,renm,gyang,yizheyang\}@bit.edu.cn}}

\begin{document}
\maketitle
\begin{abstract}
Multi-hop Question Answering (QA) requires the machine to answer complex questions by finding scattering clues and reasoning from multiple documents. Graph Network (GN) and Question Decomposition (QD) are two common approaches at present. The former uses the ``black-box'' reasoning process to capture the potential relationship between entities and sentences, thus achieving good performance. At the same time, the latter provides a clear reasoning logical route by decomposing multi-hop questions into simple single-hop sub-questions. In this paper, we propose a novel method to complete multi-hop QA from the perspective of Question Generation (QG). Specifically, we carefully design an end-to-end QG module on the basis of a classical QA module, which could help the model understand the context by asking inherently logical sub-questions, thus inheriting interpretability from the QD-based method and showing superior performance. Experiments on the HotpotQA dataset demonstrate that the effectiveness of our proposed QG module, human evaluation further clarifies its interpretability quantitatively, and thorough analysis shows that the QG module could generate better sub-questions than QD methods in terms of fluency, consistency, and diversity. 
\end{abstract}

\epigraph{\textit{"Educated mind first sign is good at asking questions."}}{--- Plekhanov G.V.}
\section{Introduction}
Unlike single-hop QA~\cite{rajpurkar-etal-2016-squad,trischler-etal-2017-newsqa,lai-etal-2017-race} where the answers could usually be derived from a single paragraph or sentence, multi-hop QA~\cite{welbl2018constructing,yang2018hotpotqa} is a challenging task that requires soliciting hidden information from scattered documents on different granularity levels and reasoning over it in an explainable way. 

\begin{figure}[t]
    \centering
    \includegraphics[width=0.4\textwidth]{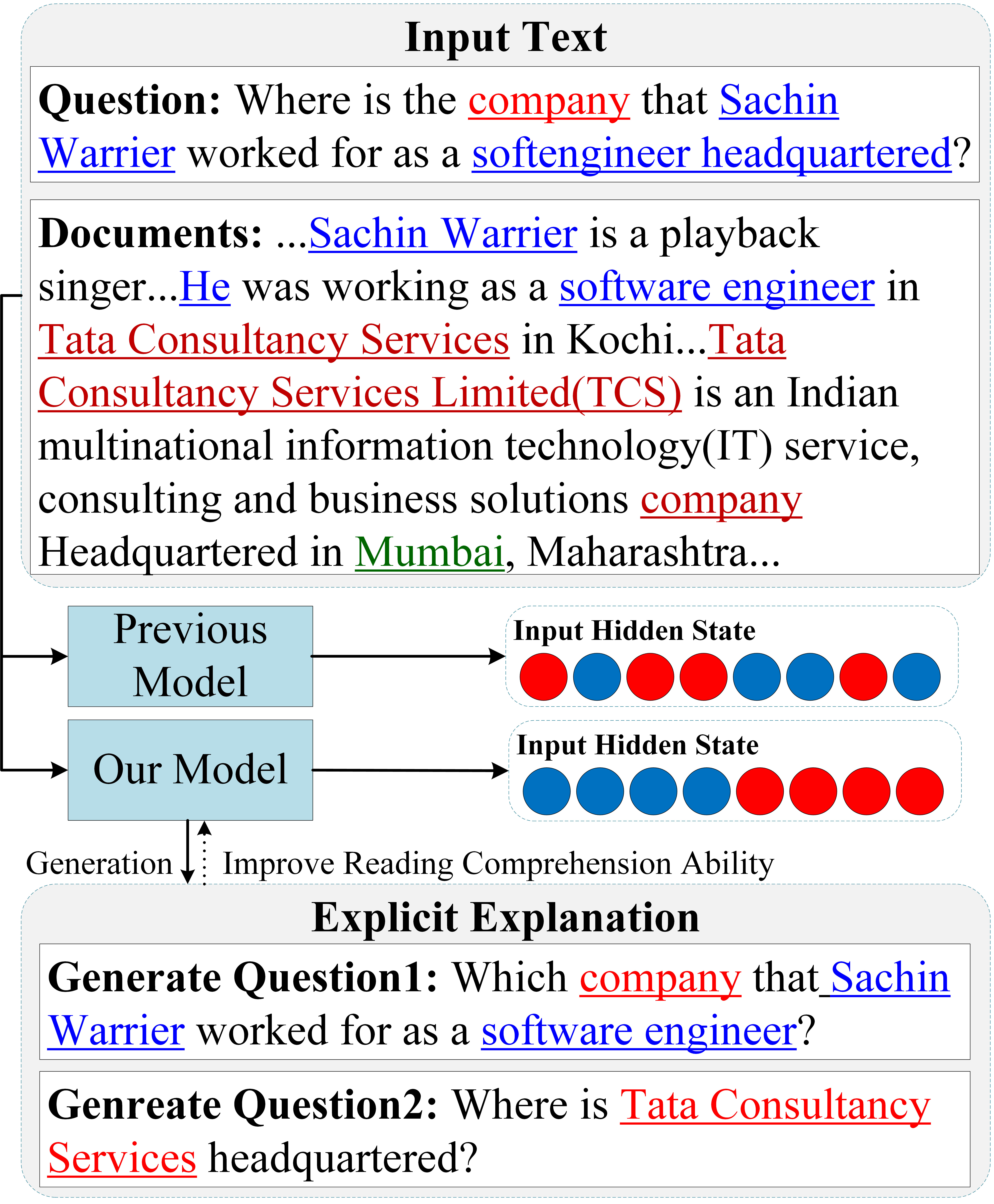}
    \caption{An example from HotpotQA dataset. Text in \textcolor{blue}{blue} is the first-hop information and text in \textcolor{red}{red} is the second-hop information. The mixed encoding of the first-hop information (\textcolor{blue}{$\bullet$}) and the second-hop information (\textcolor{red}{$\bullet$}) will confuse models with weaker reading comprehension.}
    \label{fig:example}
\end{figure}

\par
The HotpotQA~\cite{yang2018hotpotqa} was published to leverage the research attentions on reasoning processing and explainable predictions. Figure \ref{fig:example} shows an example from HotpotQA, where the question requires first finding the name of the company (\textit{Tata Consultancy Services}), and then the address of the company (\textit{Mumbai}). While, a popular stream of Graph Network-based (GN) approaches~\cite{de2019question,tu2019multi,ding2019cognitive,fang2020hierarchical} was proposed due to the structures of scattered evidence could be captured by the graphs and reflected in the representing vectors. However, the reasoning process of the GN-based method is entirely different from human thoughts. Specifically, GN tries to figure out the underlying relations between the key entities or sentences from the context. However, the process is a ``black-box''; we do not know which nodes in the network are involved in reasoning for the final answer, thus showing relatively poor interpretability.

\par
Inspired by that human solves such questions by following a transparent and explainable logical route, another popular stream of Question Decomposition-based (QD) approaches became favored in recent years~\cite{fu2021decomposing, nishida2019answering, min2019multi, jiang2019self}. The method mimics human reasoning to decompose complex questions into simpler, single-hop sub-questions; thus, the interpretability is greatly improved by exposing intermediate evidence generated by each sub-question. Nevertheless, the general performance is usually much worse than GN-based ones due to error accumulation that arose by aggregating answers from each single-hop reasoning process. Furthermore, the sub-questions are generated mainly by extracting text spans from the original question to fill the template. Hence the sub-questions are challenging to guarantee in terms of quality, such as fluency, diversity, and consistency with the original question intention, especially when the original questions are linguistically complex. 
\par
In this work, we believe that asking the question is an effective way to elicit intrinsic information in the text and is an inherent step towards understanding it~\cite{pyatkin2021asking}. Thus, we propose resolving these difficulties by introducing an additional QG task to teach the model to ask questions. Specifically, we carefully design and add one end-to-end QG module based on the classical GN-based module. Unlike the traditional QD-based methods that only rely on information brought by the question, our proposed QG module could generate fluent and inherently logical sub-questions based on the understanding of the original context and the question simultaneously. 
\par
Our method enjoys three advantages: First, it achieves better performance. Our approach preserves the GN module, which could collect information scattered throughout the documents and allows the model to understand the context in depth by asking questions. Moreover, the end-to-end training avoids the error accumulation issue; Second, it brings better interpretability because explainable evidence for its decision making could be provided in the form of sub-questions; Thirdly, the proposed QG module has better generalization capability. Theoretically, it can be plugged and played on most traditional QA models.
\par
Experimental results on the HotpotQA dataset demonstrate the effectiveness of our proposed approach. It surpasses the GN-based model and QD-based model by a large margin. Furthermore, robust performance on the noisy version of HotpotQA proves that the QG module could alleviate the shortcut issue, and visualization on sentence-level attention indicates a clear improvement in natural language understanding capability. Moreover, a human evaluation is innovatively introduced to quantify improvements in interpretability. Finally,  exploration on generated sub-questions clarifies diversity, fluency, and consistency.

\section{Related Work}

\paragraph{Multi-hop QA}
In multi-hop QA, the evidence for reasoning answers is scattered across multiple sentences. Initially, researchers still adopted the ideas of single-hop QA to solve multi-hop QA~\cite{dhingra2018neural,zhong2019coarse}. Then the graph neural network that builds graphs based on entities was introduced to multi-hop QA tasks and achieved astonishing performance~\cite{de2019question,tu2019multi,ding2019cognitive}. While, some researchers paid much attention to the interpretability of the coreference reasoning chains~\cite{fu2021decomposing, nishida2019answering, min2019multi, jiang2019self}. By providing decomposed single-hop sub-questions, the QD-based method makes the model decisions explainable.

\paragraph{Interpretability Analysis in NLP}
An increasing body of work has been devoted to interpreting neural network models in NLP in recent years. These efforts could be roughly divided into structural analyses, behavioral studies, and interactive visualization~\cite{belinkov:2019:tacl}.

Firstly, the typical way of structural analysis is to design probe classifiers to analyze model characteristics, such as syntactic structural features~\cite{elazar2021amnesic} and semantic features~\cite{wu2021infusing}. Secondly, the main idea of behavioral studies is that design experiments that allow researchers to make inferences about computed representations based on the model’s behavior, such as proposing various challenge sets that aim to cover specific, diverse phenomena, like systematicity exhaustivity~\cite{gardner2020evaluating,ravichander2021noiseqa}. Thirdly, for interactive visualization, neuron activation~\cite{durrani2020analyzing}, attention mechanisms~\cite{hao2020self} and saliency measures~\cite{janizek2021explaining} are three main standard visualization methods.

\paragraph{Question Generation}
QG is the task of generating a series of questions related to the given contextual information. Previous works on QG focus on rule-based approaches. \citet{fabbri2020templatebased} used a template-based approach to complete sentence extraction and QG in an unsupervised manner. \citet{dhole2021synqg} developed  Syn-QG using a rule-based approach. The system consists of serialized rule modules that transform input documents into QA pairs and use reverse translation counting, resulting in highly fluent and relevant results. One of the essential applications of QG is to construct pseudo-datasets for QA tasks, thereby assisting in improving their performance~\cite{zhang2019addressing, alberti2019synthetic, lee2020generating}.
\par
Our work is most related to~\citet{pyatkin2021asking}, which produces a set of questions asking about all possible semantic roles to bring the benefits of QA-based representations to traditional SRL and information extraction tasks. However, we innovatively leverage QG into complicated multi-hop QA tasks and enrich representations by asking questions at each reasoning step.

\section{Methods}

Multi-hop QA is challenging because it requires a model to aggregate scattered evidence across multiple documents to predict the correct answer.  Probably, the final answer is obtained conditioned on the first sub-question is correctly answered. Inspired by humans who always decompose complex questions into single-hop questions, our task is to automatically produce naturally-phrased sub-questions asking about every reasoning step given the original question and a passage. Following the reasoning processing, the generated sub-questions further explain why the answer is predicted.  For instance, in Figure \ref{fig:example}, the answer ``\textit{Mumbai}" is predicted to answer \textit{Question2} which is conditioned  on \textit{Question1}'s answer. More importantly, we believe that the better questions the model asks, the better it understands the reading passage and boosts the performance of the QA model in return. 
\par
\begin{figure}[t]
    \centering
    \includegraphics[width=0.4\textwidth]{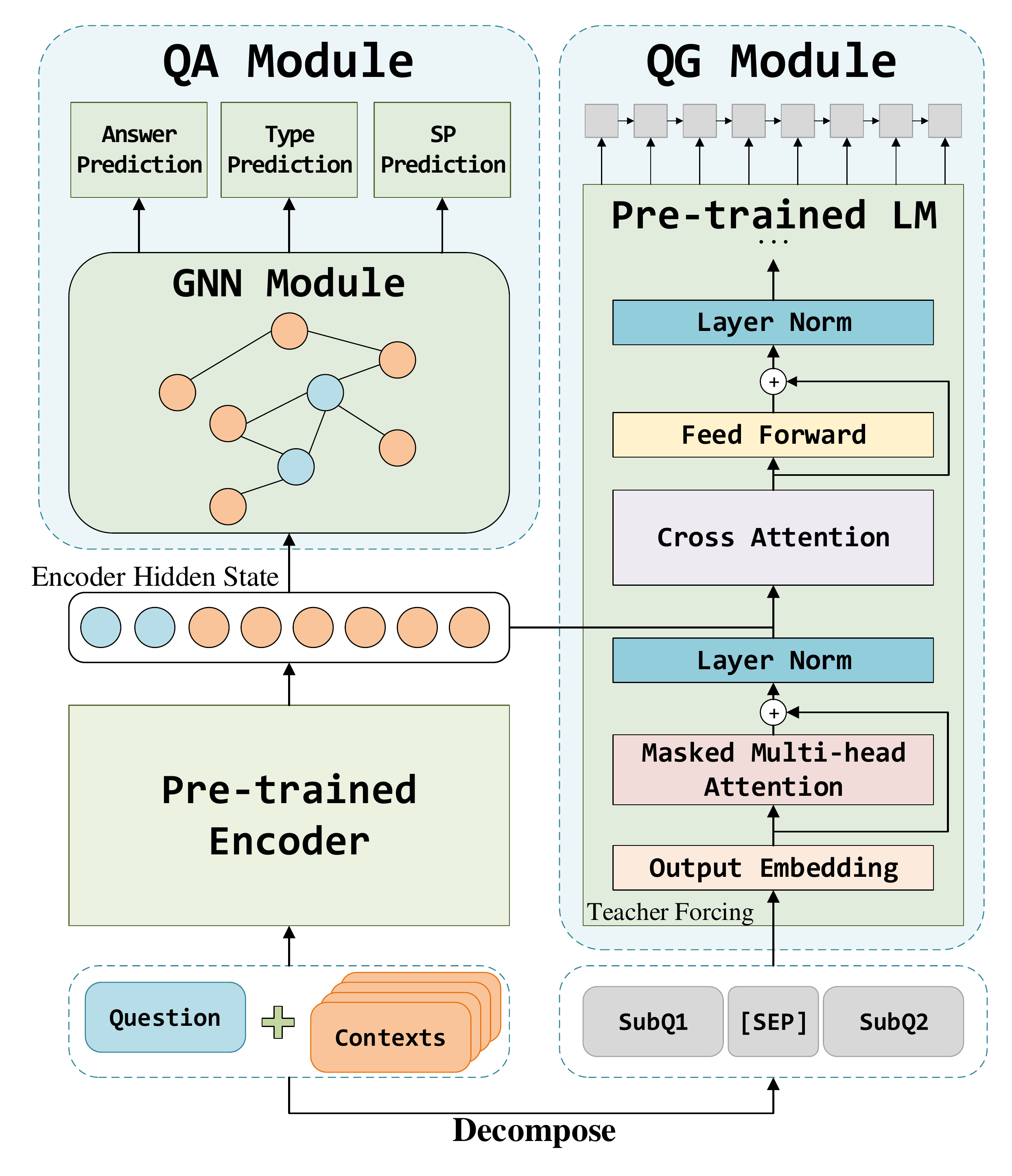}
    \caption{Overall model architecture.}
    \label{fig:model}
\end{figure}

Figure~\ref{fig:model} illustrates the overall framework of our proposed model. It consists of two modules: QA module (Section \S\ref{sec:qa}) and QG module (Section \S\ref{section:question generation}). The QA module could help model to solve multi-hop QA in a traditional way, and the QG module allows the model to solve the question in an interpretable manner by asking questions. These two modules share the same encoder and are trained end-to-end with multi-task strategy.

\subsection{Question Answering Module}
\label{sec:qa}
\paragraph{Encoder} A key point of the GN-based approach to solving QA problems is the initial encoding of entity nodes. Prior studies have shown that pre-trained models are beneficial for increasing the comprehension of the model \cite{yang-etal-2019-enhancing-pre,2020}, which enables better encoding of the input text. In Section \ref{section:question generation} we will mention that encoder will be shared to the QG module to further increase the model's reading comprehension of the input text through the QG task. Here we chose BERT as the encoder considering its strong performance and simplicity.

\paragraph{GNN Encode Module} The representation ability of the model will directly affect the performance of QA. Recent works leverage graphs to represent the relationship between entities or sentences, which have strong representation ability~\cite{xiao2019dynamically, tu2020select, fang2020hierarchical}. We believe that the advantage of graph neural networks is essential for solving multi-hop questions. Thus, we adopt the GN-based model DFGN\footnote{QA module is not the main focus of this work, and DFGN is one of the representative off-the-shelf QA models. In fact, any QA model could be adopted to replace it.}~\cite{xiao2019dynamically} that has been proven to be effective in HotpotQA.

\citet{xiao2019dynamically} build graph edges between two entities if they co-exist in one single sentence. After encoding the question $Q$ and context $C$ by the pre-trained encoder, DFGN extracts the entities' representation from the encoder output by their location information. Both mean-pooling and max-pooling are used to represent the entities' embeddings. Then, a graph neural network propagates node information to its neighbors. A soft mask mechanism is used to calculate the relevance score between each entity and the question in this process. The soft mask score is used as the weight value of each entity to indicate its importance in the graph neural network computation. 
At each step, the query embedding should be updated by the entities embedding of the current step by a bi-attention network~\cite{seo2018bidirectional}. The entities embeddings in the $t$-th reasoning step:
\begin{equation}
    \mathbf{E}^{t} = \text{GAT}([m_{1}^{t-1}e_{1}^{t-1}, m_{2}^{t-1}e_{2}^{t-1},...,m_{n}^{t-1}e_{n}^{t-1}]),
\end{equation}
where $e_{i}^{t-1}$ is the $i$-th entity's embedding at the ($t-1$)-th step and $e_{i}^{0}$ is the $i$-th entity's embedding produced both mean-pooling and max-pooling results from encoder output according to its position. $m_{i}^{t-1}$ is the relevance score, which is also called soft mask score in previous, between $i$-th entity and the question at the ($t-1$)-th step calculated by an attention network. GAT is graph attention networks proposed by~\citet{velivckovic2017graph}.

In each reasoning step, every entity node gains some information from its neighbors. An LSTM layer is then used to produce the context representation:
\begin{equation}
    \mathbf{C}^{t} = \text{LSTM}([\mathbf{C}^{t-1};\mathbf{ME}^{t\top}]),
\end{equation}
where $M$ is the adjacency matrix which records the location information of the entities.

The updated context representations are used for different sub-tasks: (i) answer type prediction; (ii) answer start position and answer end position; (iii) extract support facts prediction. All three tasks are jointly performed through multitasking learning.
\begin{equation}
    \mathcal L_{qa}=\lambda_{1}\mathcal L_{start} + \lambda_{2}\mathcal L_{end} + \lambda_{3}\mathcal L_{type} + \lambda_{4}\mathcal L_{para},
\end{equation}
where $\lambda_{1}$,$\lambda_{2}$,$\lambda_{3}$,$\lambda_{4}$ are hyper-parameters\footnote{In our experiments, we set $\lambda_{1}=\lambda_{2}=\lambda_{3}=1$,$\lambda_{4}=5$}.

\subsection{Question Generation Module}\label{section:question generation}

\paragraph{Question Generation Training Dataset} 
\label{sec:pointer}
A key challenge of training the QG module is that it is challenging to obtain the annotated sub-questions dataset. To achieve this, we take the following steps to generate sub-question dataset automatically:

First of all, according to the annotations provided by the HotpotQA dataset, the questions in the training set could be classified into the following two types: \textbf{Bridge} (70\%) and \textbf{Comparison} (30\%), where the former one requires finding evidence from first hop reasoning then use it to find second-hop evidence, while the latter requires comparing the property of two different entities mentioned in the question.

Then we leverage the methods proposed by~\citet{min2019multi} to process these two types respectively. Specifically, we adopt an off-the-shelf span predictor \texttt{\textbf{Pointer}} to map the question into several points, which could be for segmenting the question into various text spans. 

Finally, we generated sub-questions by considering the type of questions and index points provided by \texttt{\textbf{Pointer}}. Concretely, for \textbf{Bridge} questions like \textit{Kristine Moore Gebbie is a professor at a university founded in what year?}, \texttt{\textbf{Pointer}} could divided the question into two parts: \textit{Kristin Moore Gebbie be a professor at a university} and \textit{founded in what year?}. Then some question words are inserted into the first part as the first-hop evidence like \textit{Kristin Moore Gebbie be a professor at which university}, denoted as $S^{A}$. Afterward, an off-the-shelf single QA model is used to find the answer for the first sub-question, and the answer would be used to form the second sub-question like \textit{Flinders University founded in what year?}, denoted as $S^{B}$. On the other hand, for \textbf{Comparison} questions like \textit{Do The Importance of Being Icelandic and The Five Obstructions belong to different film genres?}. \texttt{\textbf{Pointer}} would divide it into three parts: first entity(\textit{The Importance of Being Icelandic}), second entity (\textit{The Five Obstructions}), and target property (\textit{film genre}). Then two sub-questions could be further generated by inserting question words to these parts like $S^{A}:$\textit{Do The Importance of Being Icelandic belong to which film genres?} and $S^{B}:$ \textit{Do The Five Obstructions belong to which film genres?}

\paragraph{Pre-trained Language Model (LM) as Generator}

After automatically creating the sub-question dataset, the next step is to train the QG module from scratch. Specifically, the structure of whole QG module is designed as seq2seq, where it shares the encoder with QA module and adopts GPT-2~\cite{radford2019language} as the decoder. During training stage, the input of decoder is formed as: $[\text{bos},y_1^A,y_2^A,...,y_n^A,\text{[SEP]}, y_1^B,y_2^B,...,y_n^B,\text{eos}]$, where [SEP] is the separator token, \textbf{bos} is the start token and \textbf{eos} is the end token.  $y_i^A$ and $y_i^B$ are the i-th token in constructed sub-questions $S^{A}$ and $S^{B}$ respectively.

Then the training objective of the QG module is to maximize the conditional probability of the target sub-questions sequence as follows:
\begin{equation}
\begin{split}
    \mathcal L_{qg} = \sum_{i=1}^{n}\text{log}\mathcal{P}(y_t|y_{<t}, h),
\end{split}
\end{equation}
where $h$ is encoder hidden state. Finally, QG module and QA module are trained together in end-to-end multi-task manner, and the overall loss is defined as $\mathcal L_{\text{multitask}}=  \mathcal L_{qa} + \mathcal L_{qg}$.

\section{Experiments}

\subsection{Dataset}\label{section:datasets}
We evaluate our approach on HotpotQA~\cite{yang2018hotpotqa} under the distraction setting, a popular multi-hop QA dataset taking the explanation ability of models into accounts. Expressly, for each question, two gold paragraphs with ground-truth answers and supporting facts are provided, along with 8 ‘distractor’ paragraphs that were collected via bi-gram TF-IDF retriever (i.e., 10 paragraphs in total). Furthermore, HotpotQA contains two types of subtasks: a) Answer prediction; and b) Supporting facts prediction; both subtasks adopt the same evaluation metrics: Exact Match (EM) and Partial Match (F1).

\subsection{Implementation Details}
We implement the model via HuggingFace library~\cite{wolf2020huggingfaces}. In detail, DFGN is selected as a QA module by following the details provided by~\cite{xiao2019dynamically}. While, for the QG module, the pre-trained decoder language model is initialized with GPT2~\cite{radford2019language}. The number of shared encoder layers is set as 12, the number of decoder layers is 6, the maximum sequence length is 512. We train the model on four TITAN RTX GPUs for 30 epochs at a batch size of 8, where each epoch tasks for around 2 hours. We select Adam~\cite{kingma2017adam} as our optimizer with a learning rate of 5e-5 and a warm-up ratio of 10\%. In general, we determine the hyperparameters by comparing the final EM and F1 scores.

\subsection{Comparison Models}
\paragraph{Baseline Model} A neural paragraph-level QA model introduced in~\citet{yang2018hotpotqa} and original proposed by~\citet{clark2018simple}.
\paragraph{DFGN}The classic GN-based  model~\cite{xiao2019dynamically} , which is trained in an end-to-end fashion for multi-hop QA task. We select this as the primary QA module in our approach, and reproduce the DFGN model by using the BERT-base pre-trained model under the hyperparameter settings released by~\citet{yang2018hotpotqa}.
\paragraph{DecompRC} The classic QD-based model that decomposes each question into several sub-questions~\cite{min2019multi}. We reproduce the DecompRC model by following the same QD instruction illustrated in~\citet{min2019multi}.

\begin{table}[t]
\resizebox{\linewidth}{!}{%
\centering
\begin{tabular}{lrrrrrr}
\hline
\multirow{2}{*}{Model} & \multicolumn{2}{c}{Answer} & \multicolumn{2}{c}{Sup Fact} & \multicolumn{2}{c}{Joint} \\ 
                       & EM           & F1          & EM            & F1           & EM          & F1          \\ \midrule
Baseline Model         & 44.44        & 58.28       & 21.95         & 66.66        & 11.56       & 40.86       \\
DecompRC               & 55.20         & 69.63       & -             & -            & -           & -           \\
DFGN* (Bridge)                   & 53.38        & 69.14       & 47.72         & \textbf{84.44 }       & 29.79       & 58.67      \\
DFGN* (Comparison)                   &\textbf{ 63.75}        & 69.48       & 70.68        & 89.98       & 46.74      & 63.56       \\
DFGN* (Total)            & 55.46    & 69.21  & 52.33  & 82.12  & 33.19  & 59.66  \\
DFGN (Total)                 & 55.66        & 69.34       & 53.10          & 82.24        & 33.68       & 59.86       \\\midrule
Ours (Bridge)                  & \textbf{56.24}        & \textbf{71.67}       & \textbf{51.06}         & 81.16        & \textbf{33.61}       & \textbf{61.75}       \\
Ours (Comparison)                  & 63.08       & \textbf{69.59}       & \textbf{73.03}         & \textbf{90.36}        & \textbf{49.23}       & \textbf{64.45}       \\
Ours (Total)                  & \textbf{57.79}        & \textbf{71.36}       & \textbf{55.77}         & \textbf{83.33}        & \textbf{36.99}       & \textbf{62.52}       \\\midrule
\end{tabular}%
}
\caption{Performance comparison on the development set of HotpotQA in the distractor setting. * indicates the results implemented by us. }
\label{table:main}
\end{table}

\section{Analysis}
Table \ref{table:main} shows the performance of different models in the development set of HotpotQA. In general, our method achieves a solid improvement on all tasks compared with either the GN-based method or QD-based one, which proves that the introduction of the QG task can effectively enhance the textual understanding ability of the model. 
Furthermore, our method achieves a consistent improvement on both types of questions. In particular, the performance on bridge-type questions requiring linear reasoning chains improves significantly, demonstrating the effectiveness of asking questions at each reasoning step.
In the following parts, we will delve into QG module functionality, interpretability, and the quality of the generated sub-questions.  
\begin{figure}[t]
    \scriptsize
    \centering
    \begin{tabularx}{\linewidth}{|c X|}
    \toprule
        \multirow{4}{*}{~} & {\bf \texttt{Question:}\,} \textcolor{red}{2014 S/S} is the debut album of a \textcolor{red}{South Korean boy group} that was formed by who?  \\
                             & {\bf \texttt{Support Fact1:}\,} \textcolor{red}{2014 S/S} is the debut album of South Korean group \textcolor{red}{WINNER}. \\
                             & {\bf \texttt{Support Fact2:}\,} Winner, often stylized as \textcolor{red}{WINNER}, is a South Korean boy group formed in 2013 by \textcolor{red}{YG Entertainment} and debuted in 2014.  \\
                             & {\bf \texttt{reasoning chain:}\,} \textcolor{red}{2014 S/S} →  \textcolor{red}{WINNER} → \textcolor{red}{YG Entertainment}\\ 
                             \midrule
        \multirow{6}{*}{~}
                             & {\bf \texttt{Noisy Fact1:}\,} Juarez, often stylized as \textcolor{blue}{Juarez}, is a South Korean boy group formed in 2013 by \textcolor{blue}{YG Arthur} and debuted in 2014. \\
                             & {\bf \texttt{Noisy Fact2:}\,} Epic, often stylized as \textcolor{blue}{Epic}, is a South Korean boy group formed in 2013 by \textcolor{blue}{YG Republic} and debuted in 2014. \\
                             & {\bf \texttt{Noisy Fact3:}\,} ...   \\
                             & {\bf \texttt{No reasoning chain with Support Fact1!}\,} \\
                             \midrule
                             & {\bf \texttt{Right Answer:}\,} \textcolor{red}{YG Entertainment} (from ours)\\ 
                             & {\bf \texttt{Disturbances:}\,} \textcolor{blue}{YG Arthur}; \textcolor{blue}{YG Republic} (from baselines) \\
    \bottomrule
    \end{tabularx}
    \caption{An example of the noisy dataset. The \textcolor{red}{red text} indicates a reasoning path with complete reasoning logic. The \textcolor{blue}{blue text} indicates some other entities which have a similar structure with the red texts, but they can be inferred from the logical relationships.}
    \label{fig:Adv-example}
\end{figure}

\subsection{Does it alleviate shortcut problem by adding question generation module?}\label{section:qg}
To verify that the QG module has the capability of focusing on discovering the real reasoning process, rather than finding shortcuts to predict answers, we further conduct QA tasks using baselines and our model on Adversarial MultiHopQA, which was first introduced in~\citet{Jiang2019reasoningshortcut}. Specifically, multiple noisy facts constructed by replacing the entities in the reasoning chain would be added to the original HotpotQA dataset to confuse the model.  For the example shown in Figure~\ref{fig:Adv-example}, the noisy facts are constructed by replacing key entities that appeared in Support Fact2. Such noisy facts have the same sentence structure as support facts but present different meanings, thus forcing the model to understand the context comprehensively. 

 Table \ref{table:shortcuts} shows the performance between the DFGN model and our model on the Adversarial-MultiHopQA dataset. In general, DFGN experiences a significant decline in performance, indicating that the existing QA model has poor robustness and is vulnerable to adversarial attacks. This further indicates that the model solves questions by mostly remembering patterns. On the other hand, by adding a QG module, the performance degradation of our method is significantly reduced. We think this is mainly because asking questions is an important strategy for guiding the model to understand the text.
 
We further prove this point through a case study shown in Figure~\ref{fig:Adv-example}. To answer the original question, the correct reasoning chain is \textit{2014 S/S} $\rightarrow$ \textit{WINNER} $\rightarrow$ \textit{YG Entertainment}. However, when there exists an overlap in the context between facts (i.e. \textit{South Korean boy group}), the current main-stream method, which strengthens representation by solely capturing internal relationships over entities or documents, usually regards the incorrect entity (i.e. \textit{YG Arthur} or \textit{YG Republic)} as a key node of reasoning chain, where so-called shortcut issue. It does not understand the reasoning process but remembers certain context patterns. However, our method mitigates such issues by reinforcing representations by asking a question at each reasoning step. As such, it could remain robust despite these disturbances. 



\begin{table}[t]
\small
\centering
\begin{tabular}{lcc}
\hline
\multicolumn{1}{c}{\multirow{2}{*}{Model}} & \multicolumn{2}{c}{Answer}         \\ \cline{2-3} 
\multicolumn{1}{c}{}                       & EM              & F1               \\ \midrule
DFGN                                       & 55.66           & 69.34            \\
DFGN*                                      & 48.08(-13.62\%) & 61.28(-11.62\%) \\ \midrule
Ours                                       & 57.79           & 71.36            \\
Ours*                                      & 52.34\textbf{(-9.43\%)}  & 65.12\textbf{(-8.74\%)}   \\ \hline
\end{tabular}
\caption{Performance of DFGN model and ours on HotpotQA dataset and its noisy version  Adversarial-MultiHopQA (marked with *).}
\label{table:shortcuts}
\end{table}

\begin{figure*}[t]
    \begin{center}
    \includegraphics[width=2.05\columnwidth]{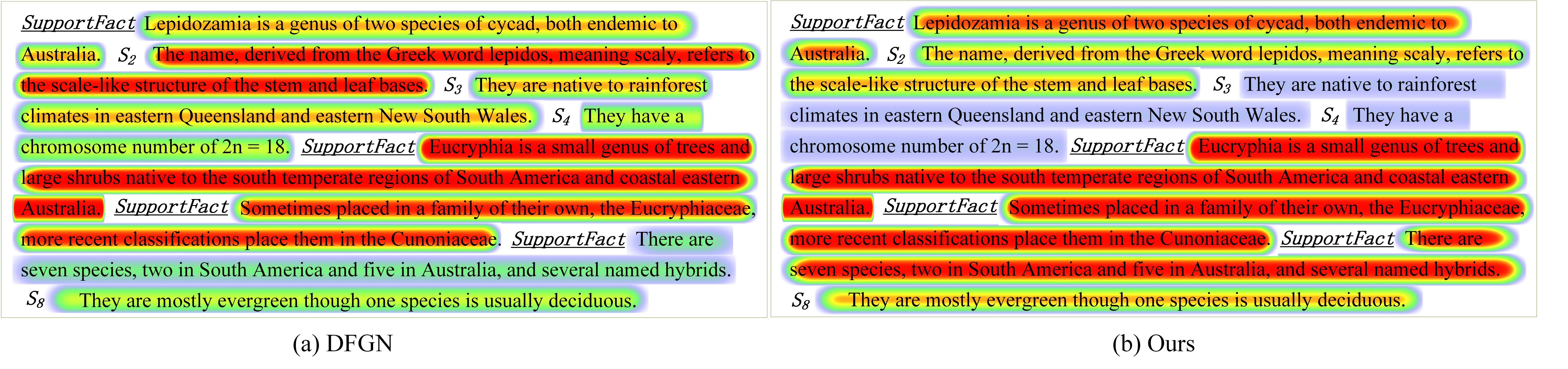}
    \end{center}
    \caption{Visualization of attention weights at sentence-level between DFGN and our method. The depth of the color corresponds to the higher attention weights of the sentence.}
    \label{fig:attention case}
\end{figure*}

\subsection{Does generated sub-question provide better interpretability?}\label{sec:interpretability}

\begin{table}[t]
\small
\centering
\begin{tabular}{ccc}
\hline
Group       & \multicolumn{1}{l}{Accuracy} & \multicolumn{1}{l}{Time(s)} \\ \hline
 A (Support Facts) & 65.63\%                      & 981                         \\
 B (Sub-questions)  & \textbf{85.94\%}                      & \textbf{543}                         \\ \hline
\end{tabular}
\caption{Average results for accuracy and time elapsed of human evaluation.}
\label{tab: interpretability}
\end{table}
Past works have proved that interpretability can be improved by exposing evidence from decomposed sub-questions. However, few quantitative analyses have been carried out on interpretability due to its subjective nature. In this paper, we use human evaluation to quantify the improvement of interpretability brought by our QG module.
\par
Specifically, we design human evaluation by following steps: First, we assemble 16 well-educated volunteers and divide them into two groups, A and B; Second, we randomly sample 8 Bridge type questions from the dev set and manually write out the correct two-hop reasoning chain for solving each question. Afterward, we replace the entity that appeared in each correct reasoning chain with other confusing entities selected from context to generate three more wrong reasoning chains (i.e., each question has 4 reasoning chains.). Then shuffle them and combine them with the original question to form a four-way multi-choice QA; Third, for each group, we ask them to figure out the correct reasoning chain and record the time elapsed for finishing all questions. To be noticed, besides original questions and reasoning chains, we provide different additional information for each group to facilitate them, all supporting facts for Group A, and all sub-questions generated by our QG for Group B. For more details, please refer to Appendix.
\par
Table~\ref{tab: interpretability} presents the results of the two groups. Remarkably, Group B has higher accuracy and takes less time. Therefore, we could argue that sub-questions generated by our QG contain more concise and precise explanations for problem-solving and further proves that the QG module can indeed improve interpretability.

\subsection{Does asking questions enhance the natural language understanding capability?}
In this work, we believe that the ability to exhaustively generate a set of logical questions according to a complex scenario allows for a comprehensive, interpretable, and flexible way of excavating the information hidden in natural language text, thereby enhancing the natural language understanding ability. 
\par
The self-attention mechanism in the pre-trained model is crucial for the model to understand the input information. Generally, the more critical a sentence is in its context, the greater attention weights it deserves. Thus, to verify whether the QG module could edify the model to carry out deep understanding intrinsically, we compare the sentence-level attention weight of our model with and without the QG module. In particular, we account for the number of increases in attention weight of support facts after adding the QG module. As shown in the last row of Table~\ref{table:characteristics}, the attention weight of around 80\% of support facts is increased, which proves that the model is more prone to focus on meaningful information with the aid of the QG task.
\par
Furthermore, Figure~\ref{fig:attention case} visualizes the changes in attention weights over supporting facts between DFGN and our method. In this case, sentences $S_{1,5,6,7}$ are considered as supporting facts. DFGN fails to predict all supporting facts and focuses on the wrong ones while our method works properly.


\begin{table}[t]
\resizebox{\linewidth}{!}{
\begin{tabular}{llccc}
\hline
Indicators       & Methods         & Win     & Tie     & Loss    \\ \hline
Diversity        & QG vs. QD & \textbf{57.64\%} & 26.70\% & 15.66\% \\
LM Score         & QG vs. QD & \textbf{60.22\%} & -       & 39.78\% \\
Attention weight & QG vs. w/o QG & \textbf{79.51\%} & -       & 20.49\% \\ \hline
\end{tabular}
}
\caption{Comparison between sub-questions generated by QG and template on diversity, LM score and Attention weights.}
\label{table:characteristics}
\end{table}

\subsection{Characteristics of Generated Questions} \label{section:pretrained LM QG}
\begin{table*}[]
\small
\resizebox{\textwidth}{!}{
\begin{tabular}{clllll}
\hline
ID                                     & \multicolumn{2}{l}{}           & Question / Sub-question                                                                                                      & Fluency            & Diversity          \\ \hline
\multicolumn{1}{c}{\multirow{5}{*}{1}} & \multicolumn{2}{l}{Question}   & In 1991 Euromarche was bought by a chain that operated how any hypermarkets at the end of 2016?             &                    &                    \\
\cline{2-6}
\multicolumn{1}{c}{}                   & \multirow{2}{*}{QD} & Q1 & Which chain that operated how any hypermarkets?                                                             & \multirow{2}{*}{×} & \multirow{2}{*}{×} \\
\multicolumn{1}{c}{}                   &                           & Q2 & In 1991 Euromarche was bought by Euromarche at the end of 2016?                                             &                    &                    \\
\cline{2-6}
\multicolumn{1}{c}{}                   & \multirow{2}{*}{QG}       & Q1 & In 1991 Euromarche was bought by which chain?                                                               & \multirow{2}{*}{\checkmark} & \multirow{2}{*}{\checkmark} \\
\multicolumn{1}{c}{}                   &                           & Q2 & Carrefour's oprated how many hypermarkets at the end of 2016?                                               &                    &                    \\ \hline
\multirow{5}{*}{2}                     & \multicolumn{2}{l}{Question}   & Do The Importance of Being Icelandic and The Five Obstructions belong to different film genres?             &                    &                    \\
                                       \cline{2-6}
                                       & \multirow{2}{*}{QD} & Q1 & Do the Importance of Being Icelandic and The Five Obstructions belong to different film genres?             & \multirow{2}{*}{×} & \multirow{2}{*}{×} \\
                                       &                           & Q2 & Do the importance of?                                                                                       &                    &                    \\
                                       \cline{2-6}
                                       & \multirow{2}{*}{QG}       & Q1 & Does the Importance of Being Icelandic and The Five Obstructions belong to which film genres?               & \multirow{2}{*}{\checkmark} & \multirow{2}{*}{\checkmark} \\
                                       &                           & Q2 & Does The Five Obstructions belong to which film genres?                                                     &                    &                    \\ \hline
\multicolumn{6}{c}{......}                                                                                                                                                                                                      \\ \hline
\multirow{5}{*}{7404}                  & \multicolumn{2}{l}{Question}   & Who was known by his stage name Aladin and helped organizations improve their performance as a consultant?  &                    &                    \\
                                       \cline{2-6}
                                       & \multirow{2}{*}{QD} & Q1 & Who was known by his stage name Aladin?                                                                     & \multirow{2}{*}{\checkmark} & \multirow{2}{*}{×} \\
                                       &                           & Q2 & Who helped organizations improve their performance as a consultant?                                         &                    &                    \\
                                       \cline{2-6}
                                       & \multirow{2}{*}{QG}       & Q1 & His stage name Aladdin?                                                                                     & \multirow{2}{*}{×} & \multirow{2}{*}{\checkmark} \\
                                       &                           & Q2 & Who was known by his stage name Aladdin and helped organizations improve their performance as a consultant? &                    &                    \\ \hline
\multirow{5}{*}{7405}                  & \multicolumn{2}{l}{Question}   & Which American film actor and dancer starred in the 1945 film Johnny Angel?                                 &                    &                    \\
                                       \cline{2-6}
                                       & \multirow{2}{*}{QD} & Q1 & Which 1945 file Johnny Angel?                                                                               & \multirow{2}{*}{×} & \multirow{2}{*}{-} \\
                                       &                           & Q2 & Which American film actor and dancer starred in noir?                                                       &                    &                    \\
                                       \cline{2-6}
                                       & \multirow{2}{*}{QG}       & Q1 & Which American file actor and dancer?                                                                       & \multirow{2}{*}{\checkmark} & \multirow{2}{*}{-} \\
                                       &                           & Q2 & Which starred in the 1945 film Johnny Angel?                                                                &                    &                    \\ \hline
\end{tabular}
}
\caption{Results on linguistic \textbf{fluency} and \textbf{diversity} of sub-questions generated by QG compared to those generated by template. \checkmark indicate the method performs better, × indicate performs worse, and - indicate performs competitively.}
\label{tab:QDvsour}
\end{table*}

QG can indeed promote an in-depth understanding of the model, but \textit{what are the characteristics of the generated questions that contribute to this?} Specifically, what are the distinctive features of the sub-questions we generate using the QG module compared to the previous QD-based methods, which generate sub-questions using templates. Through case and statistical analysis, we find that the sub-questions generated by the QG module exhibit the following characteristics:
\paragraph{Consistency} 
As mentioned in Section~\ref{sec:pointer}, previous QD-based methods require a span predictor to segment question into text spans. The errors are easy to accumulate during segmentation, and the generated sub-questions are prone to be inconsistent with the original question, especially when the original question is linguistically complex. As the second example shown in Table~\ref{tab:QDvsour}, the two sub-questions generated by template-based methods decompose the original question incorrectly, resulting in the question intention being entirely different from the original one. Therefore, such sub-questions with inconsistent intent can lead the model to misunderstand. However, our proposed QG module could carry out a comprehensive understanding of the original question with rich context information to generate sub-question in logical order. Finally, the intention of the joint sub-questions could be kept consistent with that of the original question.

\paragraph{Fluency} The fluency of a sentence could directly affect the expression of meaning, especially for questions. When a question is grammatically incorrect or incoherent, it would be difficult for people or models to understand and even misunderstand the intent of the problem. Such an issue is pervasive and inevitable in most datasets because their questions are often manually constructed, like the first example shown in Table~\ref{tab:QDvsour}. There exists a typo error (\textit{how many} $\rightarrow$ \textit{how any}) in the original question that caused the intent to change, and it is still possible to guess the correct answer from other information provided by the original question and commonsense knowledge.
Nevertheless, the sub-question generated by the QD-based method inherits the typo, and the model fails to understand correct intention due to limited information brought by the sub-question. Furthermore, the syntactic error is easy to accumulate because the boundary and attribute of the text spans are difficult to determine, resulting in poor readability.
\par
However, on the one hand, our QG module can make use of contextual information and knowledge stored in the language model to correct typo errors. On the other hand, it could take advantage of the pre-trained language model to generate fluent sentences and mitigate syntactic errors. We use Language Model Score (LMS)\footnote{https://github.com/simonepri/lm-scorer} for fluency evaluation, and as Table~\ref{table:characteristics} shows, over 60\% questions generated by QG modules have higher scores than that of the QD method.

\paragraph{Diversity} \citet{sultan-etal-2020-importance} demonstrates that the diversity of the generated question could directly influence QA performance. However, sub-questions generated by QD methods are usually monotonous and tedious due to the limitations on vocabulary and templates. While, our proposed QG module could alleviate these issues softly and enrich the diversity of questions. Depending on the pretrained LM, the QG module could copy the appropriate word in the context into sub-questions according to the different situations, such as \textit{Carrefour} shown in the first example of Table~\ref{tab:QDvsour}, which make the sub-questions more diverse and reasonable. Thus, we account for how many of the words in the sub-questions did not appear in the original question. As shown in Table~\ref{table:characteristics}, around 57\% sub-questions generated by our method are more diverse.

\section{Conclusion}
In this paper, inspired by human cognitive behavior, we believe that asking questions is an important indication to testify whether the model truly understands the input text. Therefore, we propose a QG module to solve multi-hop QA task in an interpretable manner. Based on the traditional QA module, the addition of the QG module could effectively improve the natural language understanding capability and bring superior and robust performance through asking questions. Moreover, we quantitatively analyze interpretability provided by sub-questions via human evaluation, and further clarify the interpretability via attention visualization. At last, we verify that the sub-questions obtained by the QG method are better in terms of linguistic fluency, consistency, and diversity than those obtained by the QD method.

%


\bibliography{custom}
\bibliographystyle{acl_natbib}

\appendix

\section{Appedix: Human Evaluation Instruction}\label{sec:appendix}
Specifically, we design human evaluation by following steps: 
\begin{enumerate}
    \item[1.] We assemble 16 well-educated volunteers and randomly divide them into two groups, A and B. Each group contains 8 volunteers and evenly gender.
    \item[2.] We randomly sample 8 Bridge type\footnote{Because Bride type questions always has deterministic linear reasoning chains.} questions from the dev set, and manually write out the correct two-hop reasoning chain for solving each question.
    \item[3.] We replace the entity that appeared in each correct reasoning chain with other confusing entities selected from context to generate three more wrong reasoning chains (i.e., each question has 4 reasoning chains.), then shuffle them and combine them with the original question to form a four-way multi-choice QA.
    \item[4.] For group A, except the original question, final answer and four reasoning chains, we also provide supporting facts. Then volunteers are asked to find the correct reasoning chain.
    \item[5.] For group B, except the original question, final answer and four reasoning chains, we also provide the sub-questions generated by our QG module. Then volunteers are asked to find the correct reasoning chain.
    \item[6.] We count the accuracy and time elapsed for solving problem.
\end{enumerate}

Beyond that, some details are worth noting:
\begin{enumerate}
    \item[$\bullet$] The volunteers participated in the human evaluation test are all well-educated graduate students with skilled English.
    \item[$\bullet$] We use the online questionnaire platform to design the electronic questionnaire.
    \item[$\bullet$] The questionnaire system can automatically score according to the pre-set reference answers, and count the time spent on answering the questions. 
    \item[$\bullet$] The timer starts when the volunteer clicks ``accept" button on the questionnaire, and ends when the volunteer clicks ``submit" button.
    \item[$\bullet$] Volunteers are required to answer the questionnaire without any interruption, ensuring that all time spent is for answering questions.
    \item[$\bullet$] Before starting filling the questionnaire, we provide a sample example as instruction to teach the volunteers how to find the answer.
\end{enumerate}

\begin{figure}[ht]
    \begin{center}
    \includegraphics[width=0.99\columnwidth]{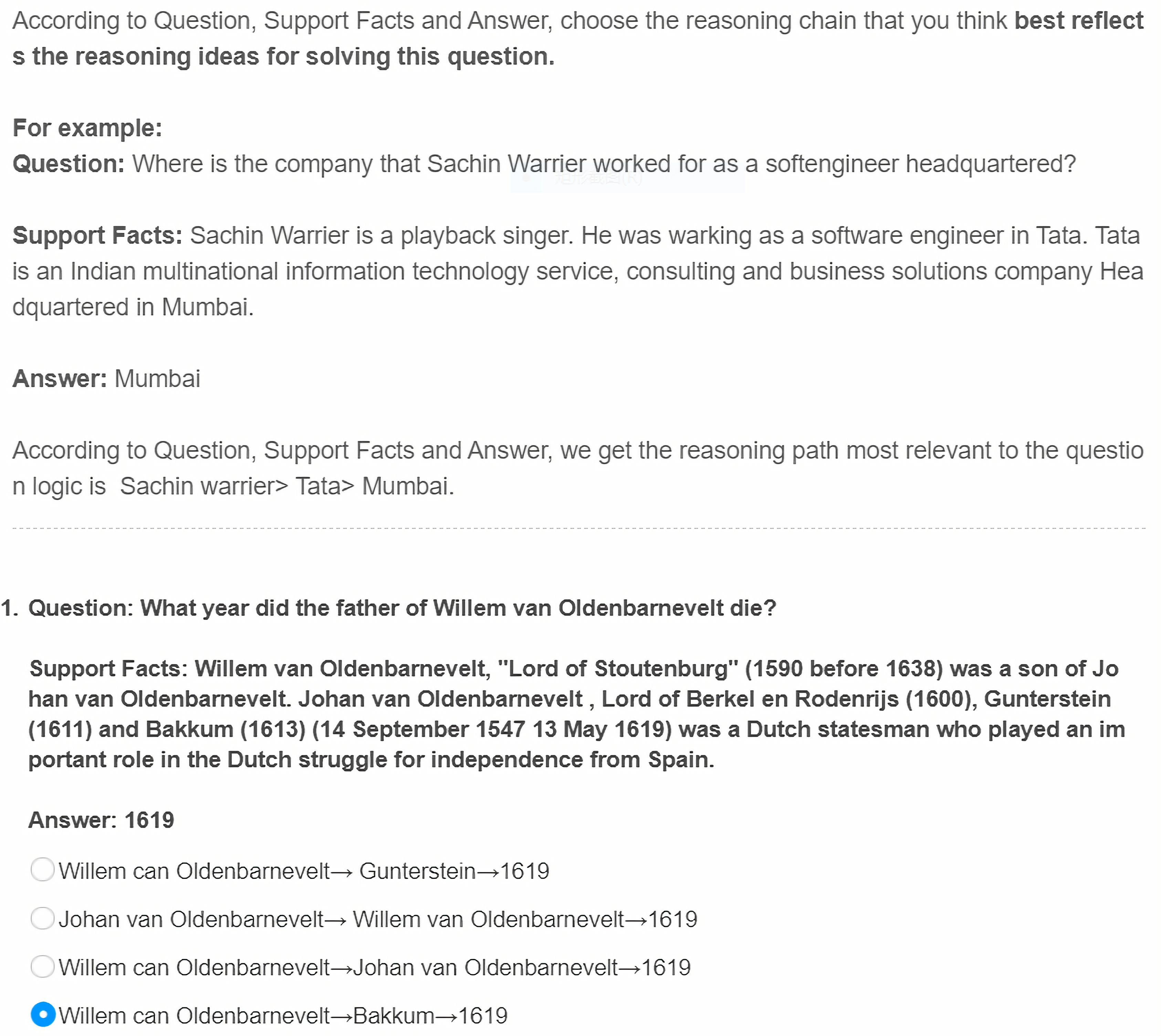}
    \includegraphics[width=0.99\columnwidth]{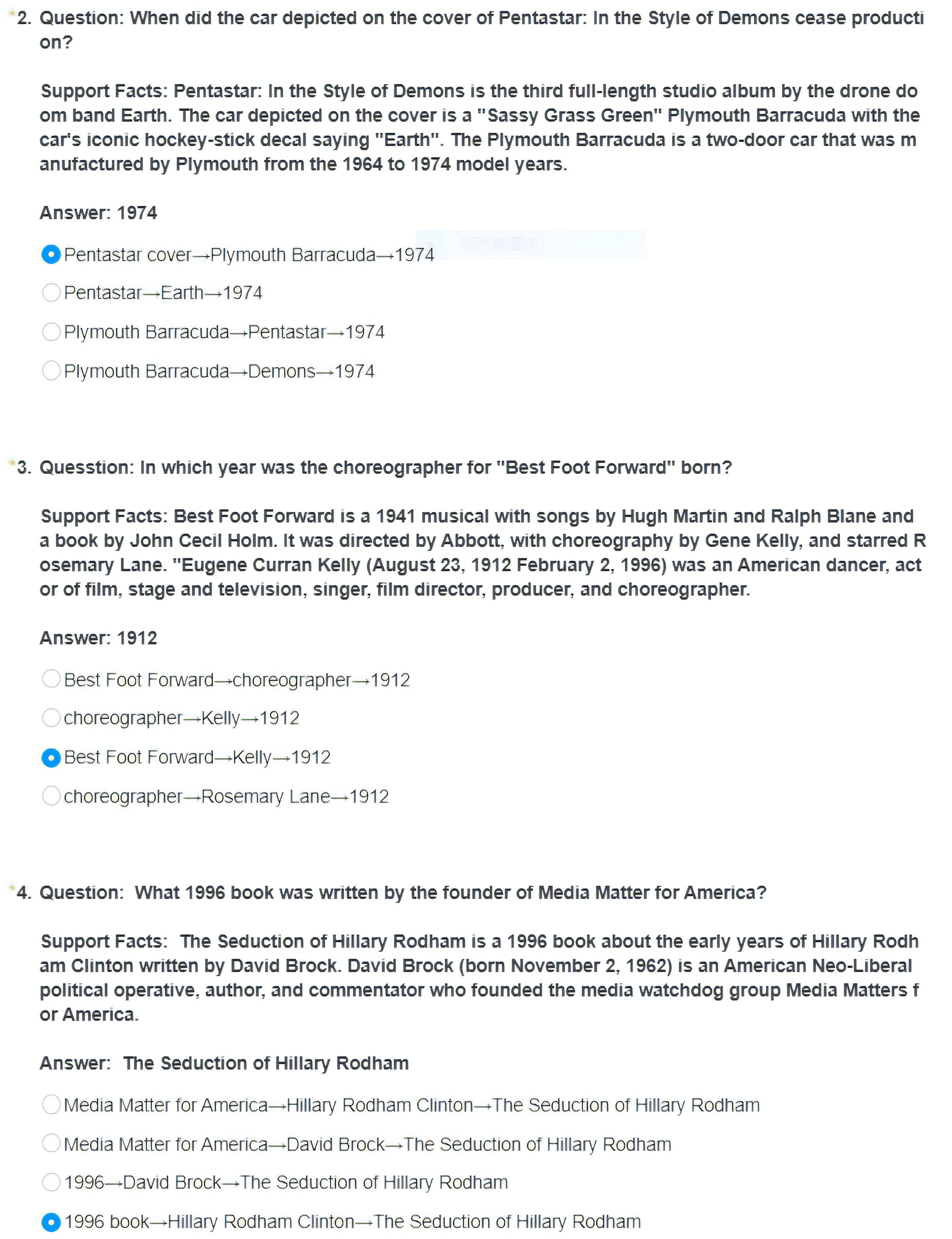}
    \end{center}
    \caption{Interface for human evaluation of choosing reasoning chain based on support facts. }
    \label{fig:sp_human}
\end{figure}

\begin{figure}[ht]
    \begin{center}
    \includegraphics[width=0.99\columnwidth]{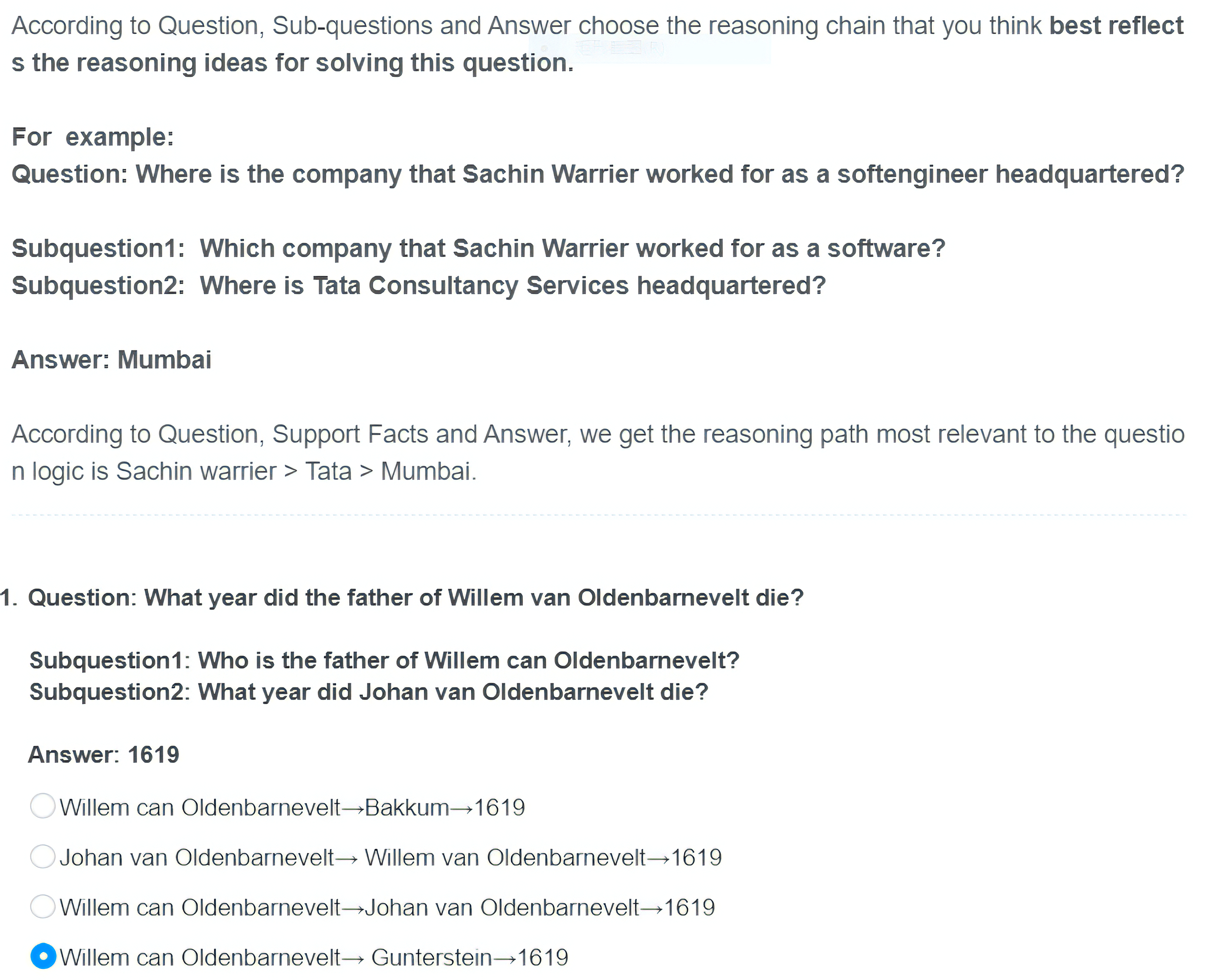}
    \includegraphics[width=0.99\columnwidth]{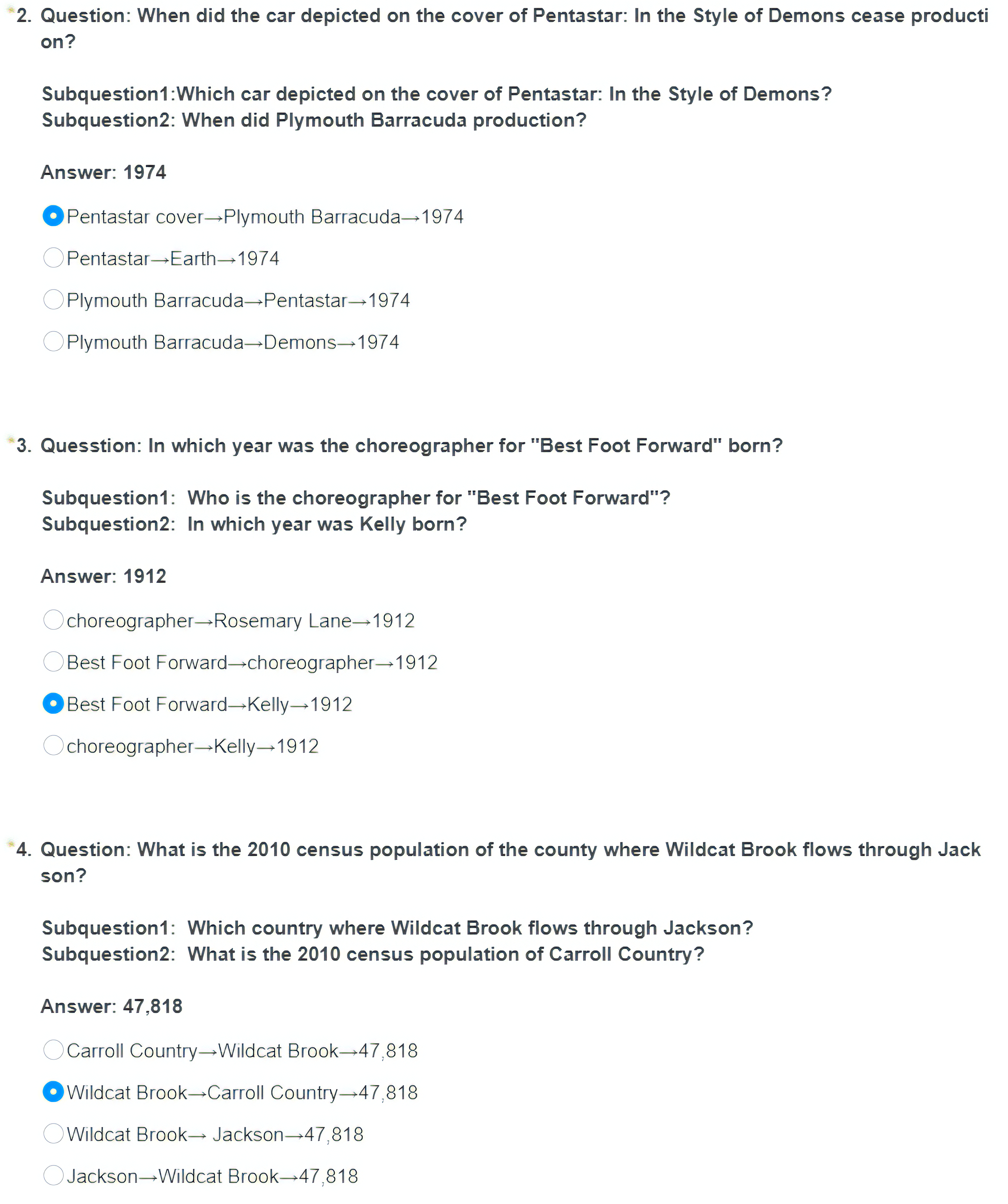}
    \end{center}
    \caption{Interface for human evaluation of choosing reasoning chain based on sub-questions.}
    \label{fig:subq_human}
\end{figure}

The interface of human evaluation for each group could be found in Figure~\ref{fig:sp_human} and Figure~\ref{fig:subq_human}.

\end{document}